%% file: main.tex
\pdfoutput=1

\documentclass[11pt]{article}

\usepackage{EMNLP2023}

\usepackage{times}
\usepackage{latexsym}

\usepackage[T1]{fontenc}

\usepackage[utf8]{inputenc}

\usepackage{microtype}

\usepackage{inconsolata}

\input{custom_commands.tex}

\usepackage{times}
\usepackage{latexsym}
\usepackage{xspace}
\usepackage{amsmath}
\usepackage{tabularx}
\usepackage{multirow}
\usepackage{booktabs}

\usepackage{xcolor}
\definecolor{dgreen}{rgb}{0,0.5,0}

\usepackage{tikz}
\pgfdeclarelayer{background}
\pgfdeclarelayer{main}
\pgfsetlayers{background,main}
\usetikzlibrary{calc}
\usepackage{ifthen}

\usepackage{graphicx}
\usepackage{caption}
\usepackage{subcaption}
\usepackage{float}
\usepackage{stfloats}
\usepackage{tikz}
\usepackage{pgfplots}
\usepgfplotslibrary{fillbetween}
\usetikzlibrary{patterns}
\pgfplotsset{compat=1.8}

\def\addlegendimage{\csname pgfplots@addlegendimage\endcsname}

\title{\gmq: Training Generalized Multi-Query Transformer Models from Multi-Head Checkpoints}

\author{
Joshua Ainslie\thanks{\- \ Equal contribution.},~ James Lee-Thorp\footnotemark[1],~ Michiel de Jong\footnotemark[1] \ \footnotemark[2]\thanks{\- \ University of Southern California. Work done at Google Research.} \\ 
{\bf Yury Zemlyanskiy},~ {\bf Federico Lebr\'{o}n},~ {\bf Sumit Sanghai}
  \AND
  {\rm \Large Google Research}\\
  }

\begin{document}
\maketitle
\begin{abstract}
Multi-query attention (\mq), which only uses a single key-value head, drastically speeds up decoder inference. However, \mq can lead to quality degradation, and moreover it may not be desirable to train a separate model just for faster inference. We (1) propose a recipe for uptraining existing multi-head language model checkpoints into models with \mq using 5\% of original pre-training compute, and (2) introduce \fullgmq attention (\gmq), a generalization of multi-query attention which uses an intermediate (more than one, less than number of query heads) number of key-value heads. We show that uptrained \gmq achieves quality close to multi-head attention with comparable speed to \mq.

\end{abstract}

\section{Introduction}
\label{section:intro}

Autoregressive decoder inference is a severe bottleneck for Transformer models due to the memory bandwidth overhead from loading decoder weights and all attention keys and values at every decoding step~\citep{shazeer2019mq, palminference, dejong2022fido}. The memory bandwidth from loading keys and values can be sharply reduced through \emph{multi-query attention}~\citep{shazeer2019mq}, which uses multiple query heads but single key and value heads.

However, multi-query attention (\mq) can lead to quality degradation and training instability, and it may not be feasible to train separate models optimized for quality and inference. Moreover, while some language models already use multi-query attention, such as PaLM~\citep{palm}, many do not, including publicly available language models such as T5~\citep{t5} and LLaMA~\citep{llama}. 

This work contains two contributions for faster inference with large language models. First, we show that language model checkpoints with multi-head attention (\mh) can be \emph{uptrained}~\citep{sparsemoe} to use \mq with a small fraction of original training compute. This presents a cost-effective method to obtain fast multi-query as well as high-quality \mh checkpoints.

Second, we propose \fullgmq attention (\gmq), an interpolation between multi-head and multi-query attention with single key and value heads \emph{per subgroup of query heads}. We show that uptrained \gmq achieves quality close to multi-head attention while being almost as fast as multi-query attention.

\section{Method}
\label{section:method}

\subsection{Uptraining}

Generating a multi-query model from a multi-head model takes place in two steps: first, converting the checkpoint, and second, additional pre-training to allow the model to adapt to its new structure. Figure \ref{fig:recycling} shows the process for converting a multi-head checkpoint into a multi-query checkpoint. The projection matrices for key and value heads are mean pooled into single projection matrices, which we find works better than selecting a single key and value head or randomly initializing new key and value heads from scratch.

\input{figures/reycling}

The converted checkpoint is then pre-trained for a small proportion $\upprop$ of its original training steps on the same pre-training recipe.

\subsection{\fullgmqcap attention}
\input{figures/gmq_architecture}
\fullgmqcap attention divides query heads into $G$ \emph{groups}, each of which shares a single key head and value head. \gmq-\textsc{g} refers to \fullgmq with $G$ groups. \gmq-$1$, with a single group and therefore single key and value head, is equivalent to \mq, while \gmq-\textsc{h}, with groups equal to number of heads, is equivalent to \mh. Figure \ref{fig:gmq_architecture} shows a comparison of \fullgmq attention and multi-head/multi-query attention. When converting a multi-head checkpoint to a \gmq checkpoint, we construct each group key and value head by mean-pooling all the original heads within that group.

An intermediate number of groups leads to an interpolated model that is higher quality than \mq but faster than \mh, and, as we will show, represents a favorable trade-off. Going from \mh to \mq reduces $H$ key and value heads to a single key and value head, reducing the size of the key-value cache and therefore amount of data that needs to be loaded by a factor of $H$. However, larger models generally scale the number of heads, such that multi-query attention represents a more aggressive cut in both memory bandwidth and capacity. \gmq lets us keep the same proportional decrease in bandwidth and capacity as model size increases.

Moreover, larger models suffer relatively less from memory bandwidth overhead from attention, as the KV-cache scales with model dimension while model FLOPs and parameters scale with the square of model dimension. Finally, standard sharding for large models replicates the single key and value head by the number of model partitions~\citep{palminference}; \gmq removes the waste from such partitioning. Therefore, we expect \gmq to present a particularly good trade-off for larger models. 

We note that \gmq is not applied to the encoder self-attention layers; encoder representations are computed in parallel, and memory bandwidth is therefore generally not the primary bottleneck.

\section{Experiments}
\label{section:experiments}

\subsection{Experimental setup}

\paragraph{Configurations}
All models are based on the T5.1.1 architecture~\citep{t5}, implemented with JAX~\citep{jax}, Flax~\citep{flax}, and Flaxformer\footnote{\url{https://github.com/google/flaxformer}}. For our main experiments we consider T5 Large and XXL with multi-head attention, as well as uptrained versions of T5 XXL with multi-query and grouped-query attention. We use the Adafactor optimizer with the same hyperparameters and learning rate schedule as T5~\citep{t5}. We apply \mq and \gmq to decoder self-attention and cross-attention, but not encoder self-attention.

\paragraph{Uptraining}

Uptrained models are initialized from public T5.1.1 checkpoints. The key and value heads are mean-pooled to the appropriate \mq or \gmq structure, and then pre-trained for a further 
$\upprop$ proportion of original pre-training steps with the original pre-training setup and dataset from~\citep{t5}. For $\upprop=0.05$, training took approximately 600 TPUv3 chip-days.

\paragraph{Data}
We evaluate on summarization datasets CNN/Daily Mail~\citep{cnn}, arXiv and PubMed~\citep{arxiv}, MediaSum~\citep{mediasum}, and Multi-News~\citep{multinews}; translation dataset WMT 2014  English-to-German; and question answering dataset TriviaQA~\citep{triviaqa}. We do not evaluate on popular classification benchmarks such as GLUE~\citep{glue} as autoregressive inference is less applicable for those tasks. 

\paragraph{Fine-tuning}

For fine-tuning, we use a constant learning rate of 0.001, batch size 128, and dropout rate 0.1 for all tasks. CNN/Daily Mail and WMT use input length of 512 and output length 256. Other summarization datasets use input length 2048 and output length 512. Finally, TriviaQA uses input length 2048 and output length 32. We train until convergence and select the checkpoint with the highest dev performance. We use greedy decoding for inference.

\paragraph{Timing}
We report time per sample per TPUv4 chip, as measured by xprof~\citep{xprof}. For timing experiments we use 8 TPUs with the largest batch size that fits up to 32 per TPU, and parallelization optimized separately for each model.
\input{tables/main_results}
\input{figures/results}
\subsection{Main results}

Figure \ref{fig:perf_vs_time} shows average performance over all datasets as a function of average inference time for \mh T5-Large and T5-XXL, and uptrained \mq and \gmq-$8$ XXL models with uptraining proportion $\upprop = 0.05$. We see that a larger uptrained \mq model provides a favorable trade-off relative to \mh models, with higher quality and faster inference than \mh-Large. Moreover, \gmq achieves significant additional quality gains, achieving performance close to \mh-XXL with speed close to \mq. Table \ref{table:headline_results} contains full results for all datasets.

\subsection{Ablations}

This section presents experiments to investigate the effect of different modeling choices. We evaluate performance on a representive subsample of tasks: CNN/Daily Mail, (short-form summarization), MultiNews (long-form summarization), and TriviaQA (question-answering).
\input{figures/conversion_methods}

\paragraph{Checkpoint conversion}
Figure \ref{fig:conversion_methods} compares the performance of different methods for checkpoint conversion. Mean pooling appears to work best, followed by selecting a single head and then random initialization. Intuitively, results are ordered by the degree to which information is preserved from the pre-trained model.

\paragraph{Uptraining steps}
Figure \ref{fig:uptraining_steps} shows how performance varies with uptraining proportion for T5 XXL with \mq and \gmq. First, we note that \gmq already achieves reasonable performance after conversion while \mq requires uptraining to be useful. Both \mq and \gmq gain from 5\% uptraining with diminishing returns from 10\%.

\input{figures/uptraining_steps}

\paragraph{Number of groups}

Figure \ref{fig:time_groups} demonstrates the effect of the number of \gmq groups on inference speed. For larger models the memory bandwidth overhead from the KV cache is less constraining~\citep{shazeer2019mq}, while the reduction in key-value size is sharper due to the increased number of heads. As a result, increasing the number of groups from \mq only results in modest slowdowns initially, with increasing cost as we move closer to \mh. We selected 8 groups as a favorable middle ground.

\input{figures/time_vs_groups}

\section{Related Work}

This work is focused on achieving a better trade-off between decoder quality and inference time through reducing the memory bandwidth overhead~\citep{roofline} from loading keys and values. \citet{shazeer2019mq} first proposed reducing this overhead through multi-query attention. Follow-up work showed that multi-query attention is especially helpful for long inputs~\citep{palminference, dejong2022fido}. \citet{gqamrabe} independently developed GQA with public implementation. Other works have explored grouping attention heads for computational efficiency~\citep{grouptrans, gmha, pillars} without focusing specifically on key-value heads, which determine memory bandwidth overhead.

A number of other methods have been proposed to reduce memory bandwidth overhead from keys and values, as well as parameters. Flash attention~\citep{flashattention} structures the attention computation to avoid materializing the quadratic attention scores, reducing memory and speeding up training. Quantization~\citep{int8, gptq} reduces the size of weights and activations, including keys and values, by lowering precision. Model distillation~\citep{distill, distillsurvey} instead reduces model size at a given precision, using data generated from the larger model to finetune the smaller model. Layer-sparse cross-attention \citep{dejong2022fido} eliminates most cross-attention layers which make up the primary expense for longer inputs. Speculative sampling~\citep{specchen, specleviathan} ameliorates the memory bandwidth bottleneck by proposing multiple tokens with a smaller model which are then scored in parallel by a larger model. 

Finally, the uptraining procedure we propose is inspired by \citet{sparsemoe}, which uptrains standard T5 checkpoints into sparsely activated Mixture-of-Experts models.

\label{section:related}

\section{Conclusion}
\label{section:conclusion}

Language models are expensive for inference primarily due to the memory bandwidth overhead from loading keys and values. Multi-query attention reduces this overhead at the cost of decreased model capacity and quality. We propose to convert multi-head attention models to multi-query models with a small fraction of original pre-training compute. Moreover, we introduce grouped-query attention, an interpolation of multi-query and multi-head attention that achieves quality close to multi-head at comparable speed to multi-query attention. 

\section*{Limitations}

This paper focuses on ameliorating the memory bandwidth overhead from loading keys and values. This overhead is most important when generating longer sequences, for which quality is inherently difficult to evaluate. For summarization we employ Rouge score, which we know is a flawed evaluation that does not tell the whole story; for that reason, it is difficult to be certain our trade-offs are correct. Due to limited computation, we also do not compare our XXL \gmq model to a comparitive model trained from scratch, so we do not know the relative performance of uptraining vs training from scratch. Finally, we evaluate the impact of uptraining and \gmq only on encoder-decoder models. Recently, decoder-only models are extremely popular, and since these models do not have separate self-attention and cross-attention, we expect \gmq to have a stronger advantage over \mq.

\section*{Acknowlegements}
We thank Santiago Onta\~{n}\'{o}n, Afroz Mohiuddin, William Cohen and others at Google Research for insightful advice and discussion. 

\bibliography{custom}
\bibliographystyle{acl_natbib}

\clearpage
\appendix

\section{Training Stability}

We find that multi-query attention can lead to training instability during fine-tuning, in particular combined with long input tasks. We trained multiple T5-Large models with multi-query attention from scratch. In each case, pre-training suffered from frequent loss spikes and the final models diverged immediately when fine-tuning on long-input tasks. Uptrained multi-query attention models are more stable but still display high variance, so for multi-query models on unstable tasks we report average performance over three fine-tuning runs. Uptrained grouped-query attention models, however, appear to be stable, so we did not investigate futher on the root causes of multi-query instability.

\end{document}

%% file: custom_commands.tex
\newcommand{\mh}{\textsc{MHA}\xspace}
\newcommand{\mq}{\textsc{MQA}\xspace}
\newcommand{\gmq}{\textsc{GQA}\xspace}
\newcommand{\fullgmq}{grouped-query\xspace}
\newcommand{\fullgmqcap}{Grouped-query\xspace}
\newcommand{\upprop}{\alpha}

\definecolor{niceblue}{rgb}{0, 0.46484375, 0.73046875} 
\definecolor{gqcolor}{rgb}{0, 0.46484375, 0.73046875} 
\definecolor{mhcolor}{rgb}{0.9296875, 0.3984375, 0.46484375} 
\definecolor{mqcolor}{rgb}{0.95, 0.52, 0.0} 
\definecolor{firstcolor}{rgb}{0.85, 0.75, 0.85} 
\definecolor{randomcolor}{rgb}{1.0, 0.7, 0.28} 
\definecolor{meancolor}{rgb}{0.21, 0.46, 0.53} 

\definecolor{darkgrey}{rgb}{0.33203125,0.33203125,0.33203125} 

%% file: figures/reycling.tex
\begin{figure}[h]
    \centering
    \includegraphics[width=0.95\columnwidth]{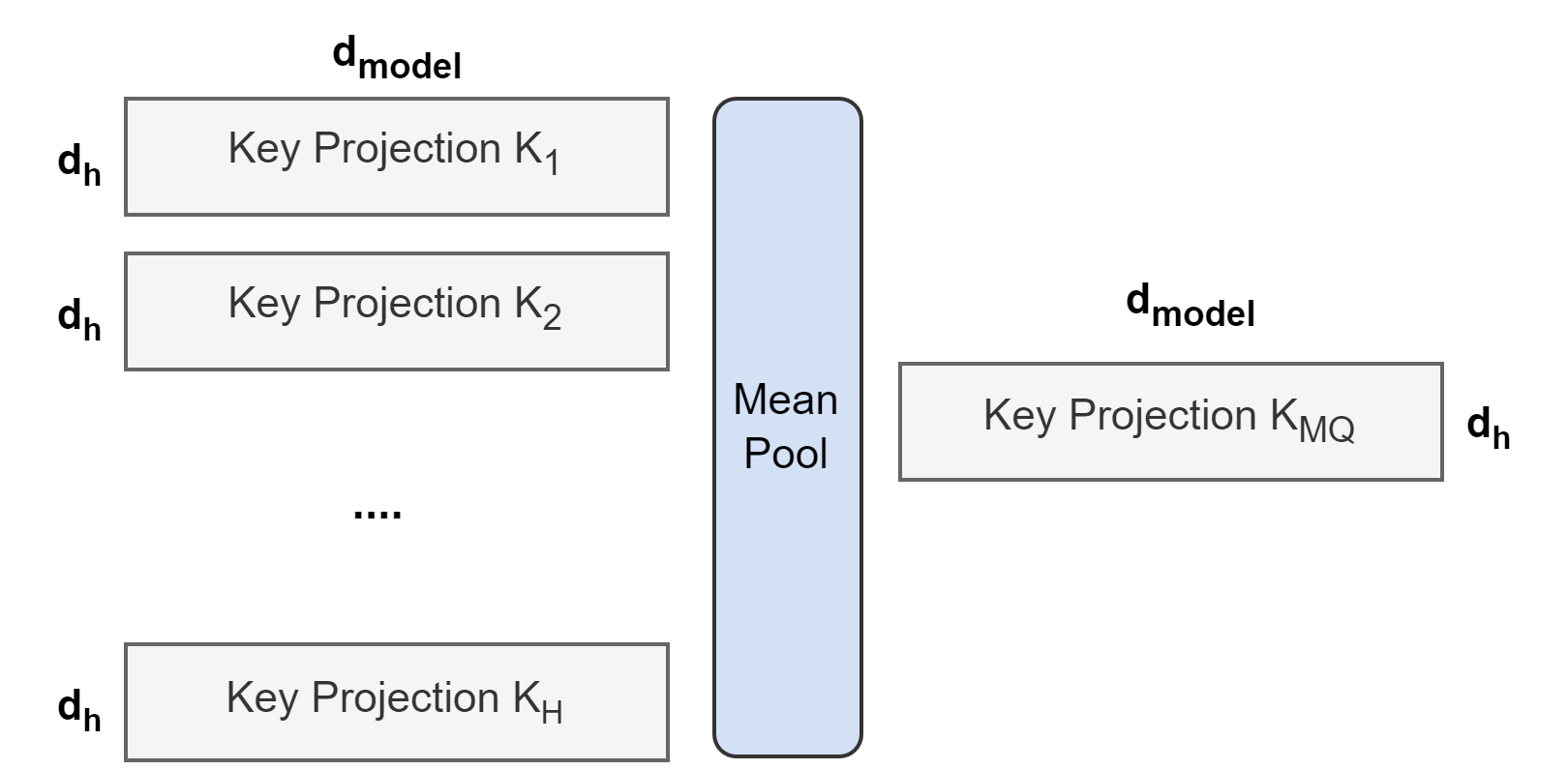}
    \caption{Overview of conversion from multi-head to multi-query attention. Key and value projection matrices from all heads are mean pooled into a single head.}
    \label{fig:recycling}
\end{figure}

%% file: figures/gmq_architecture.tex
\begin{figure*}[t]
    \centering
    \includegraphics[width=0.9\textwidth]{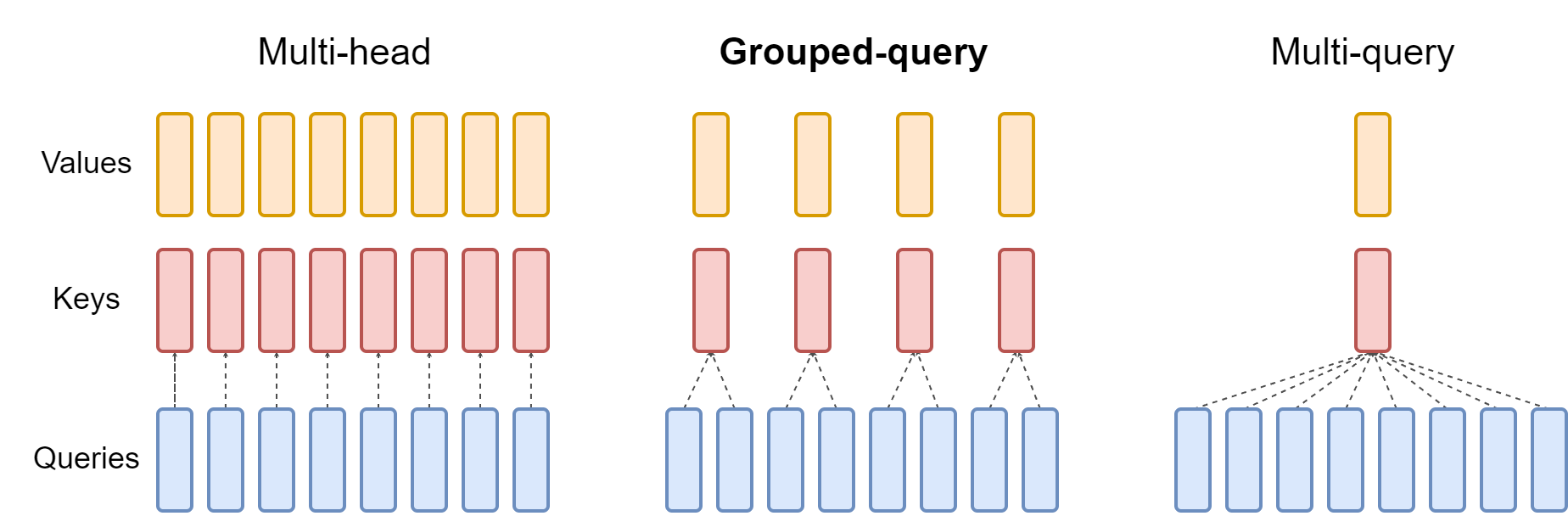}
    \caption{Overview of \fullgmq method. Multi-head attention has H query, key, and value heads. Multi-query attention shares single key and value heads across all query heads. Grouped-query attention instead shares single key and value heads for each \emph{group} of query heads, interpolating between multi-head and multi-query attention.}
    \label{fig:gmq_architecture}
\end{figure*}

%% file: tables/main_results.tex
\begin{table*}[ht!]
\centering
\footnotesize
\begin{tabular}{l|cc|ccccccccc}
\toprule
\textbf{Model} &  \textbf{T\textsubscript{infer}}& \textbf{Average} & \textbf{CNN}  &  \textbf{arXiv} & \textbf{PubMed}  &\textbf{MediaSum} & \textbf{MultiNews} &  \textbf{WMT} &  \textbf{TriviaQA}  \\
\midrule
  & \textbf{s} &  & \textbf{R\textsubscript{1}}  & \textbf{R\textsubscript{1}} &  \textbf{R\textsubscript{1}} & \textbf{R\textsubscript{1}}  &\textbf{R\textsubscript{1}} & \textbf{BLEU} & \textbf{F1}   \\
 \midrule
  \mh-Large &  0.37 & 46.0 & 42.9 & 44.6 & 46.2 & 35.5 & 46.6 & 27.7 & 78.2 \\
 \mh-XXL & 1.51 & 47.2 &  43.8 & 45.6 & 47.5 & 36.4 & 46.9 & 28.4 & 81.9 \\
 \mq-XXL  & 0.24 & 46.6 & 43.0 & 45.0 & 46.9 & 36.1 & 46.5 & 28.5 & 81.3 \\
 \gmq-8-XXL & 0.28  & 47.1 & 43.5 & 45.4 & 47.7 & 36.3 & 47.2 & 28.4 & 81.6 \\
\midrule
\bottomrule
\end{tabular}
\caption{Inference time and average dev set performance comparison of T5 Large and XXL models with multi-head attention, and 5\% uptrained T5-XXL models with multi-query and grouped-query attention on summarization datasets CNN/Daily Mail, arXiv, PubMed, MediaSum, and MultiNews, translation dataset WMT, and question-answering dataset TriviaQA.}
\label{table:headline_results}
\end{table*}

%% file: figures/results.tex
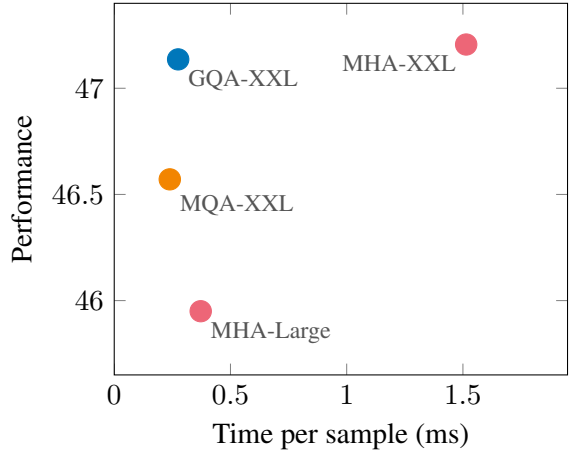
\begin{figure}[h]
\centering
\hspace{-0.5cm}
\begin{tikzpicture}
\begin{axis}[
width=0.98\columnwidth,
xmin=0.0, xmax=1.95,
ymin=45.65, ymax=47.40,
ylabel={Performance},
xlabel={Time per sample (ms)},
legend columns=1,
legend cell align=left,
legend style={
    anchor=south,
    at={(0.84, 0.05)},
},
]
\addplot[only marks, mark=*, mark options={draw=mhcolor, fill=mhcolor, scale=2}] coordinates {
    (0.372, 45.95)
    (1.514, 47.206)
};
\node at (axis cs:0.372,45.95) [anchor= north west, color=darkgrey] {\small{MHA-Large}};            
\node at (axis cs:1.514,47.206) [anchor= north east, color=darkgrey] {\small{MHA-XXL}};            
\addplot[only marks, mark=*, mark options={draw=mqcolor, fill=mqcolor, scale=2}] coordinates {
    (0.239, 46.571)
};
\node at (axis cs:0.239,46.548) [anchor= north west, color=darkgrey] {\small{MQA-XXL}};            
\addplot[only marks, mark=*, mark options={draw=gqcolor, fill=gqcolor, scale=2}] coordinates {
    (0.275, 47.136)
};
\node at (axis cs:0.275,47.136) [anchor= north west, color=darkgrey] {\small{GQA-XXL}};            
\end{axis}
\end{tikzpicture}
    \caption{{\bf Uptrained \mq yields a favorable tradeoff compared to \mh with higher quality and faster speed than \mh-Large, and \gmq achieves even better performance with similar speed gains and comparable quality to \mh-XXL.} Average performance on all tasks as a function of average inference time per sample for T5-Large and T5-XXL with multi-head attention, and 5\% uptrained T5-XXL with \mq and \gmq-8 attention.}
    \label{fig:perf_vs_time}
\end{figure}

%% file: figures/conversion_methods.tex
\begin{figure}[h]
\centering
\hspace{-0.5cm}
\begin{tikzpicture}
\begin{axis}[
    xbar,
    bar shift=0pt,
    enlarge y limits=0.3,
    width=0.95\columnwidth,
    height=0.65\columnwidth,
    bar width=16pt,
    major y tick style = transparent,    
    xmajorgrids = true,
    symbolic y coords={Mean, First, Random},
    yticklabels={Mean, First, Random},   
    ytick={Mean, First, Random},
    y dir=reverse,
    xmin=54.3,
    axis y line*=none,
    axis x line*=bottom,
]
    \addplot[style={meancolor,fill=meancolor,mark=none}]
        coordinates {(55.55,Mean)};
    \addplot[style={firstcolor,fill=firstcolor,mark=none}]
        coordinates {(55.45,First)};             
    \addplot[style={randomcolor,fill=randomcolor,mark=none}]
        coordinates {(55.23,Random)};
  
\end{axis}
\end{tikzpicture}        
    \caption{Performance comparison of different checkpoint conversion methods for T5-Large uptrained to \mq with proportion $\alpha=0.05$. `Mean' mean-pools key and value heads, `First' selects the first head and `Random' initializes heads from scratch.}
    \label{fig:conversion_methods}
\end{figure}
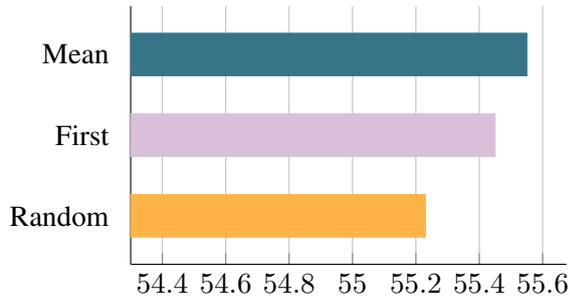

%% file: figures/uptraining_steps.tex
\begin{figure}[h]
\centering
        \begin{tikzpicture}[scale=1.0]
            \begin{axis}[
            scale only axis,
            width=0.85\columnwidth,
            height=0.45\columnwidth,
            ylabel={Performance},
            xlabel={Uptraining proportion $\upprop$},
            mark=x,
            xticklabel style={
                /pgf/number format/fixed,
                /pgf/number format/precision=2
            },            
            ymajorgrids=true,
            xmajorgrids=true,
            xminorticks=true,
            grid style=dashed,
            legend columns=1,
            legend cell align=left,
            legend style={
                anchor=south,
                at={(0.8, 0.15)},
            },
        ]
            \addplot[color=mhcolor,mark size=2pt,line width=3, dotted] table {
                0 57.537
            	0.1 57.537
            };        
            \addplot[color=gqcolor,mark=square,mark size=2pt,line width=2] table {
                0 56.713
                0.05 57.4
            	0.1 57.56
            };            
            \addplot[color=mqcolor,mark=triangle,mark size=2pt,line width=2] table {

                0 53.93
                0.05 56.92
            	0.1 57.153            	
            };
            \legend{\mh, \gmq, \mq}
            \end{axis}
        \end{tikzpicture}
    \caption{Performance as a function of uptraining proportion for T5 XXL models with \mq and \gmq-8.}
    \label{fig:uptraining_steps}
\end{figure}
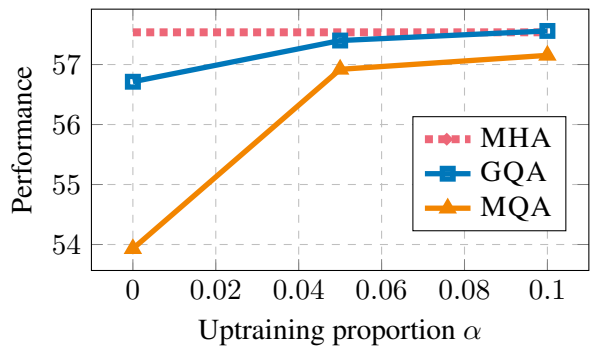

%% file: figures/time_vs_groups.tex
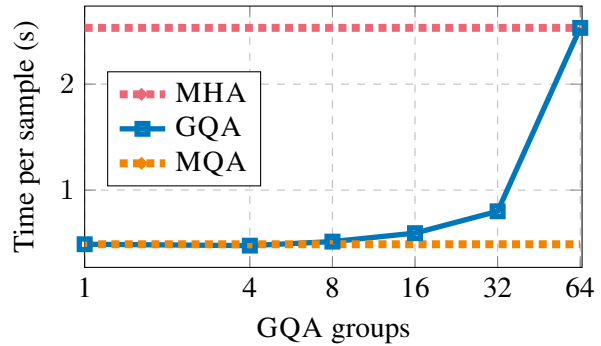
\begin{figure}[h]
\centering
        \begin{tikzpicture}[scale=1.0]
            \begin{axis}[
            scale only axis,
            xmode=log,
            width=0.85\columnwidth,
            height=0.45\columnwidth,
            ylabel={Time per sample (s)},
            xlabel={\gmq groups},
            mark=x,
            xmin=1,
            xmax=65,
            xtick={1, 4, 8, 16, 32, 64},
            xticklabels={1, 4, 8, 16, 32, 64},
            clip=false,
            ymajorgrids=true,
            xmajorgrids=true,
            xminorticks=true,
            grid style=dashed,
            legend columns=1,
            legend cell align=left,
            legend style={
                anchor=south,
                at={(0.2, 0.31)},
            },
        ]
            \addplot[color=mhcolor,mark size=2pt,line width=3, dotted] table {
                1 2.531
            	64 2.531
            };        
            \addplot[color=gqcolor,mark=square,mark size=2pt,line width=2] table {
                1 0.489
                4 0.476
                8 0.514
                16 0.594
                32 0.800
                64 2.531            

            };            
            \addplot[color=mqcolor,mark size=2pt,line width=3, dotted] table {
                1 0.489
            	64 0.489
            };
            \legend{\mh, \gmq, \mq}
            \end{axis}
        \end{tikzpicture}
    \caption{Time per sample for \gmq-XXL as a function of the number of \gmq groups with input length 2048 and output length 512. Going from 1 (\mq) to 8 groups adds modest inference overhead, with increasing cost to adding more groups.}
    \label{fig:time_groups}
\end{figure}

%% file: main.bbl
\begin{thebibliography}{28}
\expandafter\ifx\csname natexlab\endcsname\relax\def\natexlab#1{#1}\fi

\bibitem[{Bradbury et~al.(2018)Bradbury, Frostig, Hawkins, Johnson, Leary,
  Maclaurin, Necula, Paszke, Vander{P}las, Wanderman-{M}ilne, and Zhang}]{jax}
James Bradbury, Roy Frostig, Peter Hawkins, Matthew~James Johnson, Chris Leary,
  Dougal Maclaurin, George Necula, Adam Paszke, Jake Vander{P}las, Skye
  Wanderman-{M}ilne, and Qiao Zhang. 2018.
\newblock \href {http://github.com/google/jax} {{JAX}: composable
  transformations of {P}ython+{N}um{P}y programs}.

\bibitem[{Chen et~al.(2023)Chen, Borgeaud, Irving, Lespiau, Sifre, and
  Jumper}]{specchen}
Charlie Chen, Sebastian Borgeaud, Geoffrey Irving, Jean{-}Baptiste Lespiau,
  Laurent Sifre, and John Jumper. 2023.
\newblock \href {https://doi.org/10.48550/arXiv.2302.01318} {Accelerating large
  language model decoding with speculative sampling}.
\newblock \emph{CoRR}, abs/2302.01318.

\bibitem[{Chowdhery et~al.(2022)Chowdhery, Narang, Devlin, Bosma, Mishra,
  Roberts, Barham, Chung, Sutton, Gehrmann, Schuh, Shi, Tsvyashchenko, Maynez,
  Rao, Barnes, Tay, Shazeer, Prabhakaran, Reif, Du, Hutchinson, Pope, Bradbury,
  Austin, Isard, Gur-Ari, Yin, Duke, Levskaya, Ghemawat, Dev, Michalewski,
  Garcia, Misra, Robinson, Fedus, Zhou, Ippolito, Luan, Lim, Zoph, Spiridonov,
  Sepassi, Dohan, Agrawal, Omernick, Dai, Pillai, Pellat, Lewkowycz, Moreira,
  Child, Polozov, Lee, Zhou, Wang, Saeta, Diaz, Firat, Catasta, Wei,
  Meier-Hellstern, Eck, Dean, Petrov, and Fiedel}]{palm}
Aakanksha Chowdhery, Sharan Narang, Jacob Devlin, Maarten Bosma, Gaurav Mishra,
  Adam Roberts, Paul Barham, Hyung~Won Chung, Charles Sutton, Sebastian
  Gehrmann, Parker Schuh, Kensen Shi, Sasha Tsvyashchenko, Joshua Maynez,
  Abhishek Rao, Parker Barnes, Yi~Tay, Noam Shazeer, Vinodkumar Prabhakaran,
  Emily Reif, Nan Du, Ben Hutchinson, Reiner Pope, James Bradbury, Jacob
  Austin, Michael Isard, Guy Gur-Ari, Pengcheng Yin, Toju Duke, Anselm
  Levskaya, Sanjay Ghemawat, Sunipa Dev, Henryk Michalewski, Xavier Garcia,
  Vedant Misra, Kevin Robinson, Liam Fedus, Denny Zhou, Daphne Ippolito, David
  Luan, Hyeontaek Lim, Barret Zoph, Alexander Spiridonov, Ryan Sepassi, David
  Dohan, Shivani Agrawal, Mark Omernick, Andrew~M. Dai,
  Thanumalayan~Sankaranarayana Pillai, Marie Pellat, Aitor Lewkowycz, Erica
  Moreira, Rewon Child, Oleksandr Polozov, Katherine Lee, Zongwei Zhou, Xuezhi
  Wang, Brennan Saeta, Mark Diaz, Orhan Firat, Michele Catasta, Jason Wei,
  Kathy Meier-Hellstern, Douglas Eck, Jeff Dean, Slav Petrov, and Noah Fiedel.
  2022.
\newblock \href {https://doi.org/10.48550/ARXIV.2204.02311} {Palm: Scaling
  language modeling with pathways}.

\bibitem[{Cohan et~al.(2018)Cohan, Dernoncourt, Kim, Bui, Kim, Chang, and
  Goharian}]{arxiv}
Arman Cohan, Franck Dernoncourt, Doo~Soon Kim, Trung Bui, Seokhwan Kim, Walter
  Chang, and Nazli Goharian. 2018.
\newblock \href {https://doi.org/10.18653/v1/N18-2097} {A discourse-aware
  attention model for abstractive summarization of long documents}.
\newblock In \emph{Proceedings of the 2018 Conference of the North {A}merican
  Chapter of the Association for Computational Linguistics: Human Language
  Technologies, Volume 2 (Short Papers)}, pages 615--621, New Orleans,
  Louisiana. Association for Computational Linguistics.

\bibitem[{Dao et~al.(2022)Dao, Fu, Ermon, Rudra, and R{\'{e}}}]{flashattention}
Tri Dao, Daniel~Y. Fu, Stefano Ermon, Atri Rudra, and Christopher R{\'{e}}.
  2022.
\newblock \href {https://doi.org/10.48550/arXiv.2205.14135} {Flashattention:
  Fast and memory-efficient exact attention with io-awareness}.
\newblock \emph{CoRR}, abs/2205.14135.

\bibitem[{de~Jong et~al.(2022)de~Jong, Zemlyanskiy, Ainslie, FitzGerald,
  Sanghai, Sha, and Cohen}]{dejong2022fido}
Michiel de~Jong, Yury Zemlyanskiy, Joshua Ainslie, Nicholas FitzGerald, Sumit
  Sanghai, Fei Sha, and William Cohen. 2022.
\newblock \href {https://arxiv.org/abs/2212.08153} {Fi{DO}: Fusion-in-decoder
  optimized for stronger performance and faster inference}.
\newblock \emph{arXiv preprint arXiv:2212.08153}.

\bibitem[{Dettmers et~al.(2022)Dettmers, Lewis, Belkada, and
  Zettlemoyer}]{int8}
Tim Dettmers, Mike Lewis, Younes Belkada, and Luke Zettlemoyer. 2022.
\newblock \href {https://doi.org/10.48550/arXiv.2208.07339} {Llm.int8(): 8-bit
  matrix multiplication for transformers at scale}.
\newblock \emph{CoRR}, abs/2208.07339.

\bibitem[{Fabbri et~al.(2019)Fabbri, Li, She, Li, and Radev}]{multinews}
Alexander~R. Fabbri, Irene Li, Tianwei She, Suyi Li, and Dragomir~R. Radev.
  2019.
\newblock \href {https://doi.org/10.18653/v1/p19-1102} {Multi-news: {A}
  large-scale multi-document summarization dataset and abstractive hierarchical
  model}.
\newblock In \emph{Proceedings of the 57th Conference of the Association for
  Computational Linguistics, {ACL} 2019, Florence, Italy, July 28- August 2,
  2019, Volume 1: Long Papers}, pages 1074--1084. Association for Computational
  Linguistics.

\bibitem[{Frantar et~al.(2022)Frantar, Ashkboos, Hoefler, and Alistarh}]{gptq}
Elias Frantar, Saleh Ashkboos, Torsten Hoefler, and Dan Alistarh. 2022.
\newblock \href {https://doi.org/10.48550/arXiv.2210.17323} {{GPTQ:} accurate
  post-training quantization for generative pre-trained transformers}.
\newblock \emph{CoRR}, abs/2210.17323.

\bibitem[{Google(2020)}]{xprof}
Google. 2020.
\newblock {P}rofile your model with cloud tpu tools.
\newblock \url{https://cloud.google.com/tpu/docs/cloud-tpu-tools}.
\newblock Accessed: 2022-11-11.

\bibitem[{Gou et~al.(2021)Gou, Yu, Maybank, and Tao}]{distillsurvey}
Jianping Gou, Baosheng Yu, Stephen~J. Maybank, and Dacheng Tao. 2021.
\newblock \href {https://doi.org/10.1007/s11263-021-01453-z} {Knowledge
  distillation: {A} survey}.
\newblock \emph{Int. J. Comput. Vis.}, 129(6):1789--1819.

\bibitem[{Heek et~al.(2020)Heek, Levskaya, Oliver, Ritter, Rondepierre,
  Steiner, and van {Z}ee}]{flax}
Jonathan Heek, Anselm Levskaya, Avital Oliver, Marvin Ritter, Bertrand
  Rondepierre, Andreas Steiner, and Marc van {Z}ee. 2020.
\newblock \href {http://github.com/google/flax} {{F}lax: A neural network
  library and ecosystem for {JAX}}.

\bibitem[{Hinton et~al.(2015)Hinton, Vinyals, and Dean}]{distill}
Geoffrey~E. Hinton, Oriol Vinyals, and Jeffrey Dean. 2015.
\newblock \href {http://arxiv.org/abs/1503.02531} {Distilling the knowledge in
  a neural network}.
\newblock \emph{CoRR}, abs/1503.02531.

\bibitem[{Joshi et~al.(2017)Joshi, Choi, Weld, and Zettlemoyer}]{triviaqa}
Mandar Joshi, Eunsol Choi, Daniel~S. Weld, and Luke Zettlemoyer. 2017.
\newblock Triviaqa: A large scale distantly supervised challenge dataset for
  reading comprehension.
\newblock In \emph{Proceedings of the 55th Annual Meeting of the Association
  for Computational Linguistics}, Vancouver, Canada. Association for
  Computational Linguistics.

\bibitem[{Komatsuzaki et~al.(2022)Komatsuzaki, Puigcerver, Lee-Thorp, Ruiz,
  Mustafa, Ainslie, Tay, Dehghani, and Houlsby}]{sparsemoe}
Aran Komatsuzaki, Joan Puigcerver, James Lee-Thorp, Carlos~Riquelme Ruiz, Basil
  Mustafa, Joshua Ainslie, Yi~Tay, Mostafa Dehghani, and Neil Houlsby. 2022.
\newblock \href {https://doi.org/10.48550/ARXIV.2212.05055} {Sparse upcycling:
  Training mixture-of-experts from dense checkpoints}.

\bibitem[{Leviathan et~al.(2022)Leviathan, Kalman, and Matias}]{specleviathan}
Yaniv Leviathan, Matan Kalman, and Yossi Matias. 2022.
\newblock \href {https://doi.org/10.48550/arXiv.2211.17192} {Fast inference
  from transformers via speculative decoding}.
\newblock \emph{CoRR}, abs/2211.17192.

\bibitem[{Luo et~al.(2022)Luo, Zhou, Sun, Wang, Cao, Wu, Huang, and Ji}]{gmha}
Gen Luo, Yiyi Zhou, Xiaoshuai Sun, Yan Wang, Liujuan Cao, Yongjian Wu, Feiyue
  Huang, and Rongrong Ji. 2022.
\newblock \href {https://doi.org/10.1109/TIP.2021.3139234} {Towards lightweight
  transformer via group-wise transformation for vision-and-language tasks}.
\newblock \emph{{IEEE} Trans. Image Process.}, 31:3386--3398.

\bibitem[{Nallapati et~al.(2016)Nallapati, Zhou, dos Santos,
  G{\"{u}}l{\c{c}}ehre, and Xiang}]{cnn}
Ramesh Nallapati, Bowen Zhou, C{\'{\i}}cero~Nogueira dos Santos, {\c{C}}aglar
  G{\"{u}}l{\c{c}}ehre, and Bing Xiang. 2016.
\newblock \href {https://doi.org/10.18653/v1/k16-1028} {Abstractive text
  summarization using sequence-to-sequence rnns and beyond}.
\newblock In \emph{Proceedings of the 20th {SIGNLL} Conference on Computational
  Natural Language Learning, CoNLL 2016, Berlin, Germany, August 11-12, 2016},
  pages 280--290. {ACL}.

\bibitem[{Ni et~al.(2023)Ni, Mao, Yang, Lei, and Cambria}]{pillars}
Jinjie Ni, Rui Mao, Zonglin Yang, Han Lei, and Erik Cambria. 2023.
\newblock \href {https://doi.org/10.18653/V1/2023.ACL-LONG.812} {Finding the
  pillars of strength for multi-head attention}.
\newblock In \emph{Proceedings of the 61st Annual Meeting of the Association
  for Computational Linguistics (Volume 1: Long Papers), {ACL} 2023, Toronto,
  Canada, July 9-14, 2023}, pages 14526--14540. Association for Computational
  Linguistics.

\bibitem[{Park et~al.(2020)Park, Kim, Lee, Cha, Kim, and Lee}]{grouptrans}
Sungrae Park, Geewook Kim, Junyeop Lee, Junbum Cha, Ji{-}Hoon Kim, and Hwalsuk
  Lee. 2020.
\newblock \href {https://doi.org/10.18653/V1/2020.COLING-MAIN.607} {Scale down
  transformer by grouping features for a lightweight character-level language
  model}.
\newblock In \emph{Proceedings of the 28th International Conference on
  Computational Linguistics, {COLING} 2020, Barcelona, Spain (Online), December
  8-13, 2020}, pages 6883--6893. International Committee on Computational
  Linguistics.

\bibitem[{Pope et~al.(2022)Pope, Douglas, Chowdhery, Devlin, Bradbury,
  Levskaya, Heek, Xiao, Agrawal, and Dean}]{palminference}
Reiner Pope, Sholto Douglas, Aakanksha Chowdhery, Jacob Devlin, James Bradbury,
  Anselm Levskaya, Jonathan Heek, Kefan Xiao, Shivani Agrawal, and Jeff Dean.
  2022.
\newblock Efficiently scaling transformer inference.
\newblock \emph{arXiv preprint arXiv:2211.05102}.

\bibitem[{Rabe(2023)}]{gqamrabe}
Markus Rabe. 2023.
\newblock Memory-efficient attention.
\newblock
  \url{https://github.com/google/flaxformer/blob/main/flaxformer/components/attention/memory_efficient_attention.py}.
\newblock Accessed: 2023-05-23.

\bibitem[{Raffel et~al.(2020)Raffel, Shazeer, Roberts, Lee, Narang, Matena,
  Zhou, Li, and Liu}]{t5}
Colin Raffel, Noam Shazeer, Adam Roberts, Katherine Lee, Sharan Narang, Michael
  Matena, Yanqi Zhou, Wei Li, and Peter~J. Liu. 2020.
\newblock \href {http://jmlr.org/papers/v21/20-074.html} {Exploring the limits
  of transfer learning with a unified text-to-text transformer}.
\newblock \emph{J. Mach. Learn. Res.}, 21:140:1--140:67.

\bibitem[{Shazeer(2019)}]{shazeer2019mq}
Noam Shazeer. 2019.
\newblock Fast transformer decoding: One write-head is all you need.
\newblock \emph{arXiv preprint arXiv:1911.02150}.

\bibitem[{Touvron et~al.(2023)Touvron, Lavril, Izacard, Martinet, Lachaux,
  Lacroix, Rozière, Goyal, Hambro, Azhar, Rodriguez, Joulin, Grave, and
  Lample}]{llama}
Hugo Touvron, Thibaut Lavril, Gautier Izacard, Xavier Martinet, Marie-Anne
  Lachaux, Timothée Lacroix, Baptiste Rozière, Naman Goyal, Eric Hambro,
  Faisal Azhar, Aurelien Rodriguez, Armand Joulin, Edouard Grave, and Guillaume
  Lample. 2023.
\newblock \href {https://doi.org/10.48550/ARXIV.2302.13971} {Llama: Open and
  efficient foundation language models}.

\bibitem[{Wang et~al.(2019)Wang, Singh, Michael, Hill, Levy, and Bowman}]{glue}
Alex Wang, Amanpreet Singh, Julian Michael, Felix Hill, Omer Levy, and
  Samuel~R. Bowman. 2019.
\newblock \href {https://openreview.net/forum?id=rJ4km2R5t7} {{GLUE:} {A}
  multi-task benchmark and analysis platform for natural language
  understanding}.
\newblock In \emph{7th International Conference on Learning Representations,
  {ICLR} 2019, New Orleans, LA, USA, May 6-9, 2019}. OpenReview.net.

\bibitem[{Williams et~al.(2009)Williams, Waterman, and Patterson}]{roofline}
Samuel Williams, Andrew Waterman, and David~A. Patterson. 2009.
\newblock \href {https://doi.org/10.1145/1498765.1498785} {Roofline: an
  insightful visual performance model for multicore architectures}.
\newblock \emph{Commun. {ACM}}, 52(4):65--76.

\bibitem[{Zhu et~al.(2021)Zhu, Liu, Mei, and Zeng}]{mediasum}
Chenguang Zhu, Yang Liu, Jie Mei, and Michael Zeng. 2021.
\newblock \href {https://doi.org/10.18653/v1/2021.naacl-main.474} {Mediasum:
  {A} large-scale media interview dataset for dialogue summarization}.
\newblock In \emph{Proceedings of the 2021 Conference of the North American
  Chapter of the Association for Computational Linguistics: Human Language
  Technologies, {NAACL-HLT} 2021, Online, June 6-11, 2021}, pages 5927--5934.
  Association for Computational Linguistics.

\end{thebibliography}
